\documentclass[journal]{IEEEtran}

\usepackage[table]{xcolor} 
\usepackage{amsmath, amssymb, amsfonts, amsthm}
\usepackage{dsfont}
\usepackage{bm}
\usepackage{pifont}
\usepackage{bbding}

\usepackage{graphicx}
\usepackage{epstopdf}
\usepackage{float, stfloats}
\usepackage{array, multirow, booktabs, makecell, diagbox}
\usepackage{lscape}
\usepackage{caption, subcaption}
\usepackage{graphbox}
\usepackage{setspace}
\usepackage{enumerate}
\usepackage{fancyhdr}
\usepackage[normalem]{ulem}
\usepackage{soul}

\usepackage{cite}
\usepackage{algpseudocode}
\usepackage[linesnumbered,ruled]{algorithm2e}

\usepackage{hyperref}
\hypersetup{hidelinks}

\allowdisplaybreaks[4]

\newtheorem{lemma}{Lemma}

\newtheorem{proposition}{Proposition}

\newtheorem{corollary}{Corollary}

\newtheorem{property}{Property}

\newtheorem{remark}{Remark}

\newtheorem{claim}{Claim}

\title{\huge 
\texttt{LAGS}: Low-Altitude Gaussian Splatting\\with Groupwise Heterogeneous Graph Learning
}

\author{Yikun Wang,~\emph{Graduate Student Member, IEEE}, Yujie Wan,~\emph{Student Member, IEEE}, \\ Wei Zuo,~\emph{Graduate Student Member, IEEE}, Shuai Wang,~\emph{Senior Member, IEEE},\\ Yik-Chung Wu,~\emph{Senior Member, IEEE}, Chengzhong Xu,~\emph{Fellow, IEEE}, and Huseyin Arslan,~\emph{Fellow, IEEE}

\thanks{
Copyright (c) 2015 IEEE. Personal use of this material is permitted. However, permission to use this material for any other purposes must be obtained from the IEEE by sending a request to pubs-permissions@ieee.org.
}
\thanks{This work was supported by the CAS-TUBITAK Joint Call under the International Partnership Program of the Chinese Academy of Sciences (Grant No. 321GJHZ2025118M) and The Scientific and Technological Research Council of Türkiye (Grant No. 225N080), the Shenzhen Science and Technology Program (Grant No. RCYX20231211090206005), and by MOST and The Science and Technology Development Fund (FDCT), Macau SAR (File no. 0074/2025/AMJ).}
\thanks{
Yikun Wang, Wei Zuo, and Yik-Chung Wu are with The University of Hong Kong, Hong Kong, China.
Yujie Wan and Shuai Wang are with the Shenzhen Institutes of Advanced Technology, Chinese Academy of Sciences, Shenzhen, China.
Yujie Wan is also with the Southern University of Science and Technology, Shenzhen, China.
Chengzhong Xu is with the State Key Laboratory of IOTSC, University of Macau, Macau, China.
Huseyin Arslan is with College of Engineering, Istanbul Medipol University, Istanbul, Turkey. 

Yikun Wang and Yujie Wan contribute equally.

Corresponding author: Shuai Wang ({\tt\footnotesize s.wang@siat.ac.cn}).
}
}
\begin{document}

\maketitle
\begin{abstract}

Low-altitude Gaussian splatting (LAGS) facilitates 3D scene reconstruction by aggregating aerial images from distributed drones. However, as LAGS prioritizes maximizing reconstruction quality over communication throughput, existing low-altitude resource allocation schemes become inefficient. This inefficiency stems from their failure to account for image diversity introduced by varying viewpoints. To fill this gap, we propose a groupwise heterogeneous graph neural network (GW-HGNN) for LAGS resource allocation. GW-HGNN  explicitly models the non-uniform contribution of different image groups to the reconstruction process, thus automatically balancing data fidelity and transmission cost. 
The key insight of GW-HGNN is to transform LAGS losses and communication constraints into graph learning costs for dual-level message passing. Experiments on real-world LAGS datasets demonstrate that GW-HGNN significantly outperforms state-of-the-art benchmarks across key rendering metrics, including PSNR, SSIM, and LPIPS. Furthermore, GW-HGNN reduces computational latency by approximately 100x compared to the widely-used MOSEK solver, achieving millisecond-level inference suitable for real-time deployment.

\end{abstract}


\section{Introduction}

Gaussian splatting (GS) \cite{kerbl20233d} has emerged as an enabling technology for 3D reconstruction and world models \cite{lu2025gwm}. 
Conventional GS methods adopt smart phones \cite{kerbl20233d} or ground robots \cite{liu2026communication} to collect vision data. 
However, as scenes expand, these methods are constrained by limited viewpoints due to 2D camera mobility.
To address this limitation, this paper studies low-altitude GS (LAGS), which builds GS models over low-altitude wireless networks (LAWNs).
By leveraging the 3D mobility of drones in LAWNs \cite{yuan2025ground,wang2026low}, LAGS achieves higher diversity of viewpoints. 
Furthermore, since drones exist for a long period of time in a city-wide range, perceiving buildings, roads, and objects during their flights (e.g., delivery), it becomes a ``free lunch'' to achieve data scale-up by aggregating images from their historical observations \cite{turki2022mega,Li2026STTGS}.

Nonetheless, sending vast volumes of images from drones to the ground leads to a heavy communication burden. Moreover, a significant portion of these images can be redundant and of limited value for building the LAGS model.
Unfortunately, currently there is no method to select the most GS-valuable images for transmission over the LAWNs. 
Existing LAWN resource allocation approaches often optimize generic objectives, including sensing accuracy \cite{yuan2025ground}, communication throughput \cite{ye2025integrated}, or computation efficiency \cite{wang2026low}, which overlook the non-uniform contribution of different images to the reconstruction process.
While recent work~\cite{Li2026STTGS} takes a step forward by adopting a GS-specific objective to distinguish between different drones, it still treats all images from a single drone as equally important. As such, this method fails to account for the significant image diversity introduced by varying viewpoints of the same drone.

To bridge this gap, we propose a groupwise heterogeneous graph neural network (GW-HGNN) for LAGS reconstruction. 
Our method begins by partitioning the local dataset of each drone into distinct image groups based on viewpoint similarity. 
It then employs heterogeneous node types and dedicated update mechanisms to model both \emph{inter-drone and intra-drone relationships}. Crucially, by formulating GS losses and communication constraints as graph learning costs for dual-level message passing, GW-HGNN learns to jointly optimize reconstruction quality and communication efficiency through adaptive image selection and power control. {Different from generic learning-based resource allocation methods that mainly optimize communication-centric metrics such as sum rate and energy efficiency~\cite{Guo2022learning,shen2023Graph,chen2026CAPA}, GW-HGNN learns a GS-specific resource allocation policy that depends on both the communication environment and the GS reconstruction objective. Besides, although task-oriented communications has been studied for generic tasks such as classification and detection~\cite{shi2023task,Shao2022taskoriented}, existing methods do not address the LAGS-specific challenge. To the best of our knowledge, this work presents the \emph{first unified graph learning framework that seamlessly integrates GS with LAWN}.}

We conducted extensive experiments on real-world datasets~\cite{turki2022mega}. 
Results demonstrate that GW-HGNN significantly outperforms state-of-the-art baselines STT-GS \cite{Li2026STTGS}, MaxLAWN \cite{ye2025integrated}, ActiveGS \cite{jin2025activegs} across all rendering quality metrics (i.e., SSIM, PSNR, LPIPS). Furthermore, GW-HGNN achieves millisecond-level inference latency, representing a $100$x speedup (i.e., $1$\% of the runtime) compared to the commercial MOSEK solver. Ablation studies further validate the importance of our cross-layer optimization, showing consistent performance gains over alternative groupwise (GW) transmission strategies that optimize for communication or GS objectives alone.

\section{System Model} \label{section2}

We consider a LAGS system consisting of an $N$-antenna ground server and $K$ single-antenna drones. 
The ground server possesses an initial GS model $\mathcal{S}^{'}$ that involves discrepancies compared to the actual 3D scene $\mathcal{S}^{*}$.
The system aims to output a 3D GS model $\mathcal{S}$ such that $\mathcal{S}$ is closer to $\mathcal{S}^{*}$, by aggregating local datasets $\{\mathcal{D}_{k}\}$ from the drones, where each $\mathcal{D}_{k}$ with $k\in\mathcal{K}\triangleq\{1,\cdots,K\}$ contains $L_k$ pairs of camera images and view poses \cite{kerbl20233d}. 

To accomplish the above task, the data needs to be transmitted to the ground server over the LAWN. 
The received signal at the server is $\mathbf{r} = \sum_{k=1}^K \mathbf{h}_{k} z_{k}+\mathbf{n}$,
where $\mathbf{h}_{k}\in\mathbb{C}^{N}$ denotes the channel of drone $k$, $z_{k}\in\mathbb{C}$ is the transmit signal with power $p_{k}$, and $\mathbf{n}\in\mathbb{C}^{N}$ is additive white Gaussian noise with power $\sigma^2$. 
To decode individual data from each drone, we apply a receiver $\mathbf{w}_k$ to $\mathbf{r}$ \cite{Li2026STTGS,liu2026communication}. 
The uplink data rate of drone $k$ is
\begin{align}
&R_{k}=B\mathrm{log}_2\left(1+\frac{H_{kk}p_{k}}{\sum_{j=1,j\neq k}^KH_{kj}p_{j}+
\sigma^2} \right), \label{Rk}
\end{align}
where $B$ is the bandwidth of the system, and $H_{kj}$ represents the composite channel gain, defined as $H_{kk}=\|\mathbf{w}_k^H\mathbf{h}_{k}\|^{2}$ for the desired link and $H_{kj}=\|\mathbf{w}_k^H\mathbf{h}_{j}\|^{2}$ with $k\ne j$, for the interference link from drone $j$ to $k$.
If $N\gg K$, we can adopt a low complexity maximum ratio combining (MRC) receiver $\mathbf{w}_k=
\left\Vert\mathbf{h}_{k}\right\Vert_2^{-1}
\mathbf{h}_{k}$.
However, if $N$ has a moderate size, we need to adopt the more powerful interference rejection combining (IRC) receiver to suppress the interference leakage, which is 
the solution to the generalized eigenvalue problem $\mathbf{w}_k = \mathop{\arg\max}_{\mathbf{w}} \frac{\mathbf{w}^H \mathbf{h}_{k}\mathbf{h}_{k}^H \mathbf{w}}{\mathbf{w}^H 
(\sum_{j\neq k}
\mathbf{h}_{j}\mathbf{h}_{j}^H+\sigma^2\mathbf{I}_N
) \mathbf{w}}${, i.e., $\mathbf{w}_k = \mathbf{R}_{\text{int}} ^{-1}   \mathbf{h}_k / \|\mathbf{R}_{\text{int}} ^{-1}   \mathbf{h}_k \|_2$, where $\mathbf{R}_{\text{int}}=\sum_{j \neq k} p_j \mathbf{h}_j \mathbf{h}_j^H + \sigma^2 \mathbf{I}_N$~\cite{ma2017low}}.

\section{Groupwise LAGS Transmission}
\begin{figure}[!t] 
    \centering
    \setlength{\tabcolsep}{2pt}  
    \begin{tabular}{ccc}
    \includegraphics[width=0.49\textwidth]{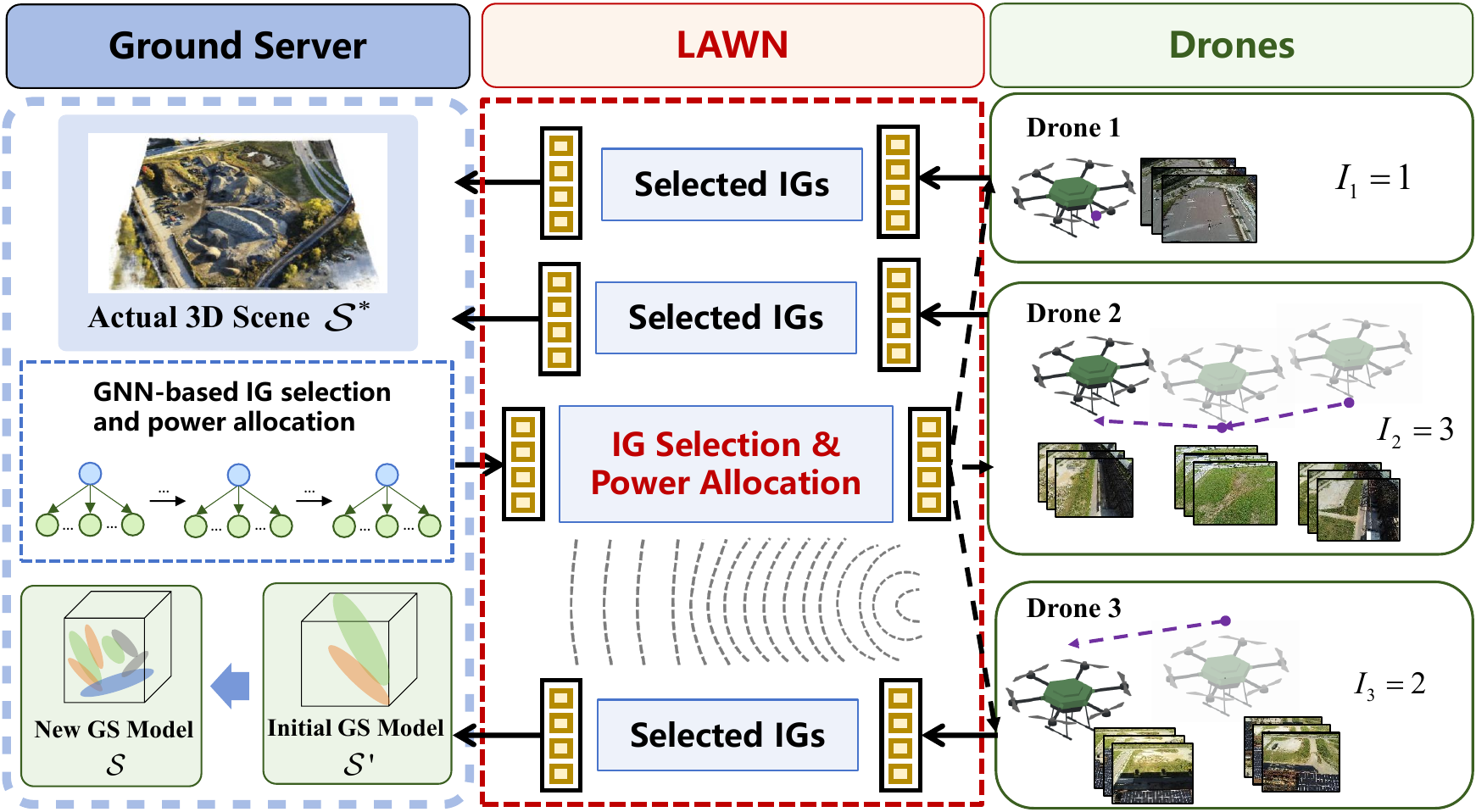} \\
    \end{tabular}
    \caption{{Overview of the proposed LAGS framework.}}
    \label{fig:VisualGroupwise}
\end{figure}

The major issue in LAGS systems is that the data volume of $\mathcal{D}_{k}$ can be large, e.g., in the range of Gigabytes or more.
Observing that a significant portion of $\mathcal{D}_{k}$ is of limited value to $\mathcal{S}$, we propose a groupwise LAGS transmission strategy, which partitions $\mathcal{D}_{k}$ into $I_{k}$ groups $\{\mathcal{D}_{ki}\}_{i=1}^{I_k}$ based on viewpoint {grouping} and only transmits a subset of $\mathcal{D}_{k}$. 
As such, the communication overhead is significantly reduced. The overall of our proposed groupwise transmission is illustrated in Fig.~\ref{fig:VisualGroupwise}.

The groupwise strategy hinges on selecting groups that simultaneously maximize reconstruction quality and communication efficiency.
Let $\mathbf{x}\in\{0,1\}^{\hat{I}}$ denote the group selection vector, where $x_{ki}=1$ indicates that group $\mathcal{D}_{ki}$ is scheduled for upload, and $\hat{I}=\sum_{k=1}^{K}I_{k}$ is the total number of groups.
The transmission of selected groups must be completed within a time budget $T$, i.e., $R_{k}^{-1}(\sum_{i=1}^{I_{k}}x_{ki}Q_{ki}) \leq T$, where $Q_{ki}$ (in bits) is the data volume of group $\mathcal{D}_{ki}$. 


Our objective is to maximize the quality of GS model $\mathcal{S}$, which is measured by the GS loss function \cite{kerbl20233d}.
Specifically, denote $\mathcal{D}_{ki} = \left\{\mathbf{v}_{l,ki}, \mathbf{s}_{l,ki}\right\}_{l=1}^{L_{ki}}$, where $\mathbf{v}_{l,ki}$ is an image, $\mathbf{s}_{l,ki}$ is its pose \cite{kerbl20233d}, and $L_{ki}=|\mathcal{D}_{ki}|$ is the number of images. 
The GS model takes a pose $\mathbf{s}_{l,ki}$ as input and renders an image using model $\mathcal{S}$ as $\hat{\mathbf{v}}_{l,ki}=\mathcal{R}(\mathbf{s}_{l,ki}|\mathcal{S})$.
The GS loss is computed as the difference between rendered and actual images:
\begin{align}
\mathcal{L}_{\text{GS}}\left(\hat{\mathbf{v}}_{l,ki},\mathbf{v}_{l,ki}\right)
=&(1-\lambda)\|\hat{\mathbf{v}}_{l,ki}-\mathbf{v}_{l,ki}\|_1
\nonumber\\
&+\lambda [1-\mathsf{SSIM}(\hat{\mathbf{v}}_{l,ki},\mathbf{v}_{l,ki})],
\end{align}
where the weight $\lambda=0.2$ according to \cite{kerbl20233d} and 
$\mathsf{SSIM}$ is the SSIM function detailed in \cite[Eqn. 5]{wang2011ssim}.
Ideally, maximizing the GS quality is equivalent to minimizing the GS loss $\sum_{\mathbf{v}_{l,ki}\in\Omega}\mathcal{L}_{\text{GS}}\left(\hat{\mathbf{v}}_{l,ki},\mathbf{v}_{l,ki}\right)$ w.r.t. $\mathcal{S}$ over full image space $\Omega$.
However, we have no access to $\mathcal{S},\Omega$ prior to LAGS transmission.
Yet, we do have access to $\mathcal{S}'$, and by the theory of uncertainty sampling~\cite{yoo2019learning, jin2025activegs}, we only need to select drone data that can change the current model $\mathcal{S}^{'}$ to the maximum extent.
This converts GS loss minimization over $\mathcal{S}$ into GS loss maximization over $\mathcal{S}'$, resulting in the following problem:
\begin{subequations}
  \label{eq:OP_and_constraints}
  \begin{align}
    \mathsf{P}: \max_{\substack{\mathbf{x},\mathbf{p}}} 
      \quad & \sum_{k=1}^{K}\sum_{i=1}^{I_{k}}x_{ki}\pi_{ki}\left(\mathcal{S}',\mathcal{D}_{ki}\right)
      \label{eq:op_with_prediction_loss} \\
    \text{s.t.}\quad & 
      TR_{k}(\mathbf{p}) \ge \sum_{i=1}^{I_{k}}x_{ki}Q_{ki}, ~ \forall k,
    \label{eq:load1_constraints} \\ 
    &p_{k}\ge0,~\forall k,~~\sum_{k=1}^K p_k  \leq P_{\text{sum}},
      \label{eq:power_limits} \\
    & x_{ki} \in \{0, 1\}, ~ \forall k,i,
      \label{eq:binary_constraint} 
  \end{align}
\end{subequations}
where $\mathbf{p}\triangleq[p_1,p_2,\dots,p_{K}]^{T}$ and satisfies $\sum_{k=1}^Kp_k\leq P_{\mathrm{sum}}$.
On the other hand, $\pi_{ki}(\mathcal{S}',\mathcal{D}_{ki})$ can be estimated by a sample-then-transmit mechanism \cite{Li2026STTGS}, which is given by
\begin{equation}
\pi_{ki}\left(\mathcal{S}',\mathcal{D}_{ki}\right)=\sum_{\left(\mathbf{v}_{l,ki},\mathbf{s}_{l,ki}\right)\in\mathcal{D}_{ki}}\mathcal{L}_{\text{GS}}\left(\mathcal{R}(\mathbf{s}_{l,ki}|\mathcal{S}'),\mathbf{v}_{l,ki}\right).
\end{equation}

\section{GW-HGNN for LAGS Reconstruction}

Problem $\sf{P}$ is a mixed integer nonconvex optimization program.
A traditional way is to convert $\sf{P}$ into a sequence of surrogate convex problems and then adopt conic solvers to solve the surrogate instances \cite{liu2026communication,Li2026STTGS,diamond2016cvxpy}. 
However, their computation complexity is {superlinear} in $\sum_{k=1}^KI_k$, which becomes prohibitively slow under the groupwise LAGS setting with $\sum_{k=1}^KI_k \gg 1$.
To address this challenge, we propose a GW-HGNN method with a much lower complexity.

\subsection{Architecture Design}

\begin{figure*}[!t] 
    \centering
    \setlength{\tabcolsep}{2pt}  
    \begin{tabular}{ccc}
    \includegraphics[clip,width=1\textwidth]{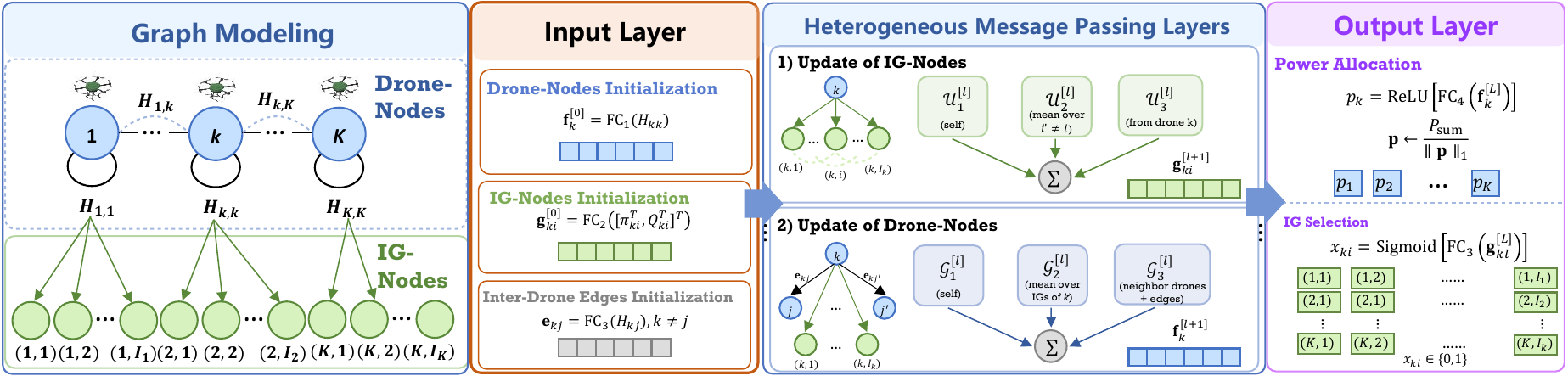} \\
    \end{tabular}
    \caption{{Architecture of the proposed GW-HGNN.}}
    \label{fig:VisualGNN}
\end{figure*}

To exploit the inherent topology structure of the LAGS system, we propose an HGNN architecture that consists of two types of nodes, i.e., drone nodes and image group (IG) nodes, as illustrated in Fig.~\ref{fig:VisualGNN}.
Compared to generic DNNs, the proposed HGNN offers two key advantages. First, its architecture is inherently scalable: the number of trainable parameters remains fixed regardless of the number of drones or IGs, enabling seamless deployment in large-scale LAGS systems. Second, HGNN respects the 2D permutation equivariance properties of $\sf{P}$, i.e., its output is consistent under arbitrary permutations of drones and groups, thereby significantly improving learning efficiency~\cite{Guo2022learning}.

The graph contains $K$ drone-nodes and $\sum_{k=1}^{K}I_{k}$ IG-nodes, reflecting the hierarchical relationship between drones and their viewpoint-based IGs. Each drone-node $k$ is initialized with its effective channel gain $H_{k,k}$, while inter-drone interference channels $H_{k,j}~(k\ne j)$ are encoded as edge features between drone-nodes $k$ and $j$. Each IG-node $(k,i)$ is initialized using the tuple $\left[\pi_{ki}, Q_{ki}\right]$, which captures the reconstruction utility and data volume of group $\mathcal{D}_{ki}$. Formally, the initial node and edge features are constructed as:
\begin{equation}
\begin{aligned}
&\mathbf{f}^{[0]}_{k}=\text{FC}_1(H_{k,k}), \quad \forall k \in \mathcal{K}, \\
&\mathbf{g}^{[0]}_{ki}=\text{FC}_2\left(\left[\pi_{ki}(\mathcal{S}',\mathcal{D}_{ki}),{Q_{ki}}\right]^T\right), \quad \forall k,i, \\
&\mathbf{e}_{kj}=\text{FC}_3(H_{k,j}), \quad \forall k,j\in\mathcal{K},~k\ne j,
\end{aligned}
\end{equation}
where $\text{FC}_{l}(\cdot)$ denotes a learnable fully connected layer. 

\begin{figure*}[!b]
\normalsize
\hrulefill
\begin{equation}
\mathcal{L}_{\text{LD}} = \mathbb{E}\left\{ - \sum_{k=1}^{K}\sum_{i=1}^{I_{k}}x_{ki}\pi_{ki}\left(\mathcal{S}',\mathcal{D}_{ki}\right) + \sum_{k=1}^{K} \mu_{k} \cdot \left[ \sum_{i=1}^{I_{k}} x_{ki} Q_{ki} - T R_{k} \right]+\psi \sum_{k=1}^{K} \sum_{i=1}^{I_{k}} \log x_{ki} \log(1 - x_{ki}) \right\}.
\label{Equ:Total_Loss_PDL}\tag{9}
\end{equation}
\end{figure*}

The HGNN then performs heterogeneous message passing to capture both \emph{intra-drone} and \emph{inter-drone} relationships. First, IG-node features are updated via intra-drone aggregation:
\begin{align}
\mathbf{g}_{ki}^{[l]}=&\mathcal{U}_{1}^{[l]}\left(\mathbf{g}_{ki}^{[l-1]}\right)+\frac{1}{I_k-1}\sum_{i'\ne i}\mathcal{U}_{2}^{[l]}\left(\mathbf{g}_{ki'}^{[l-1]}\right)
\nonumber\\
&+\mathcal{U}_{3}^{[l]}\left(\mathbf{f}_{k}^{[l-1]}\right), ~\forall k, i,
\tag{6}
\end{align}
where $\mathcal{U}^{[l]}_{1}(\cdot)$-$\mathcal{U}^{[l]}_{3}(\cdot)$ are multi-layer perceptrons (MLPs). Subsequently, drone-node features are refined by aggregating messages from their associated IG-nodes, neighboring drones, and edge features:
\begin{equation}
\begin{aligned}
\mathbf{f}_{k}^{[l]}=& \mathcal{G}_{1}^{[l]}\left(\mathbf{f}_{k}^{[l-1]}\right)+\frac{1}{I_{k}}\sum_{i}\mathcal{G}_{2}^{[l]}\left(\mathbf{g}_{ki}^{[l-1]}\right) \\
 & +\frac{1}{K-1}\sum_{j\ne k}\mathcal{G}_{3}^{[l]}\left(\mathbf{f}_{j}^{[l-1]},\mathbf{e}_{kj}\right), ~\forall k\in\mathcal{K},
\end{aligned}\tag{7}
\end{equation}
with $\mathcal{G}_{1}^{[l]}(\cdot)$-$\mathcal{G}^{[l]}_{3}(\cdot)$ being additional MLPs. 

Finally, after $L$ layers' message passing, the scheduling decisions are generated through dedicated output layers:
\begin{equation}
\begin{aligned}
&p_{k}=\text{ReLU}\left[\text{FC}_{4}\left(\mathbf{f}^{[L]}_{k}\right)\right], \forall k, ~  \mathbf{p}\gets \frac{P_{\text{sum}}}{\|\mathbf{p}\|_1} \\
& x_{ki}= \text{Sigmoid}\left[\text{FC}_{5}\left(\mathbf{g}_{ki}^{[L]}\right)\right], ~\forall k,i,
\end{aligned}\tag{8}
\end{equation}
where the raw power outputs are normalized to satisfy the total power constraint~\eqref{eq:power_limits}, while the group selection scores are mapped to the interval $(0,1)$ via Sigmoid activation.

{It is worth noting that all $\text{FC}$ layers and $\text{MLP}$ modules in the proposed GW-HGNN are shared across nodes or edges of the same type, e.g., the same drone-node update functions $\mathcal{G}_{1}^{[l]}(\cdot)$-$\mathcal{G}^{[l]}_{3}(\cdot)$ are applied to all drone-nodes. Therefore, when the number of drones or IGs increases, only the number of message-passing operations increases, while the number of trainable parameters remains fixed. This weight-sharing mechanism ensures that the number of trainable parameters remains unchanged when $K$ or $I_k$ increases.}

\subsection{Training and Inference}

Given the combinatorial and non-convex nature of the original problem, we adopt an unsupervised learning approach wherein the GW-HGNN is trained directly using the system’s objective as the loss function. A critical challenge lies in enforcing the communication load constraints in~\eqref{eq:load1_constraints}. While penalty-based methods are commonly used, they require careful tuning of penalty coefficients that are often instance-dependent and sensitive to hyperparameters.

\textit{1) Loss Function}: We employ a Lagrangian duality approach \cite{Li2024PenaltyDualLearning}, training the GNN to minimize the augmented Lagrangian of $\sf{P}$. The resultant loss is shown in~\eqref{Equ:Total_Loss_PDL}, where $\mu_{k}\ge0$ denotes the Lagrange multiplier for drone $k$, and $\psi>0$ is a fixed regularization coefficient. The first two terms together constitute the partial Lagrangian of the problem, while the last term serves as an entropy-based regularization that encourages binary decisions by penalizing values deviating from 0 or 1. The expectation is taken over the joint distribution of channel realizations and image data, and is approximated in practice by the empirical average over mini-batches during stochastic gradient descent. 

\textit{2) Training}: The Lagrange multipliers are updated via the subgradient method:
\begin{equation}
\mu_{k}\gets\mu_{k}+{\tau}\mathbb{E}\left\{\sum_{i=1}^{I_{k}} x_{ki}{Q_{ki}}-TR_{k}\right\},
\label{equ: Lagrangian Dual Approach}
\end{equation}
where $\tau>0$ denotes the step size. This update dynamically balances utility maximization and constraint satisfaction without requiring manual hyperparameter tuning. The training procedure is summarized in Algorithm~\ref{Algorithm: Lagrange Dual Learning}, where $\boldsymbol{\theta}_{\text{G}}$ denotes the GW-HGNN parameters trained with Adam optimizer.

\begin{algorithm}[t]
\caption{\texttt{GW-HGNN for LAGS}}
\label{Algorithm: Lagrange Dual Learning}
\textbf{Configure} epoch counts $N^{\text{e}}$ and steps per epoch $N^{\text{s}}$;
\\
\textbf{Initialize} $\boldsymbol{\theta}_{\text{G}}$ and $\boldsymbol{\mu}\triangleq[\mu_1,\mu_2,\dots,\mu_{K}]^{T}$\;
\For{epoch $=1,2,\dots,N^{\text{e}}$}{
    \For{step $=1,2,\dots,N^{\text{s}}$}{
        Sample a batch of system input states $\{\pi_{ki},Q_{ki},H_{kj}\}$\;
        
        Compute $\mathbf{x}$ and $\mathbf{p}$ using HGNN with parameters $\boldsymbol{\theta}_{\text{G}}$\;
        
        $\boldsymbol{\theta}_{\text{G}} \gets \text{Adam}(\boldsymbol{\theta}_{\text{G}}, \nabla_{\boldsymbol{\theta}_{\text{G}}} \mathcal{L}_{\text{LD}})$\;
        \For{$k=1,2,\dots,K$}{update $\mu_{k} $ using~\eqref{equ: Lagrangian Dual Approach};}
    }
}
\KwOut{Trained GW-HGNN with parameters $\boldsymbol{\theta}_{\text{G}}$.}
\end{algorithm}

\textit{3) Inference}: {During online inference, the trained GNN directly predicts the $\mathbf{x}$ and $\mathbf{p}$ from the inputs $\{H_{k,j}\}$, $\{\pi_{ki}\}$ and $\left\{Q_{ki}\right\}$. The continuous output of IG selection score in (9) is binarized with threshold $\eta=0.5$. Then, we check the communication load constraints in (3b). If a drone violates its constraint, i.e., the total data size of selected IGs exceed the achievable communication capacity, we iteratively remove one selected IG from that drone until the constraint is satisfied. In each step, the selected IG with the smallest GS utility $\pi_{ki}$ is removed.}

\begin{figure*}[t]
    \centering
    \begin{subfigure}[b]{0.4\linewidth}
        \centering
        \includegraphics[width=\linewidth]{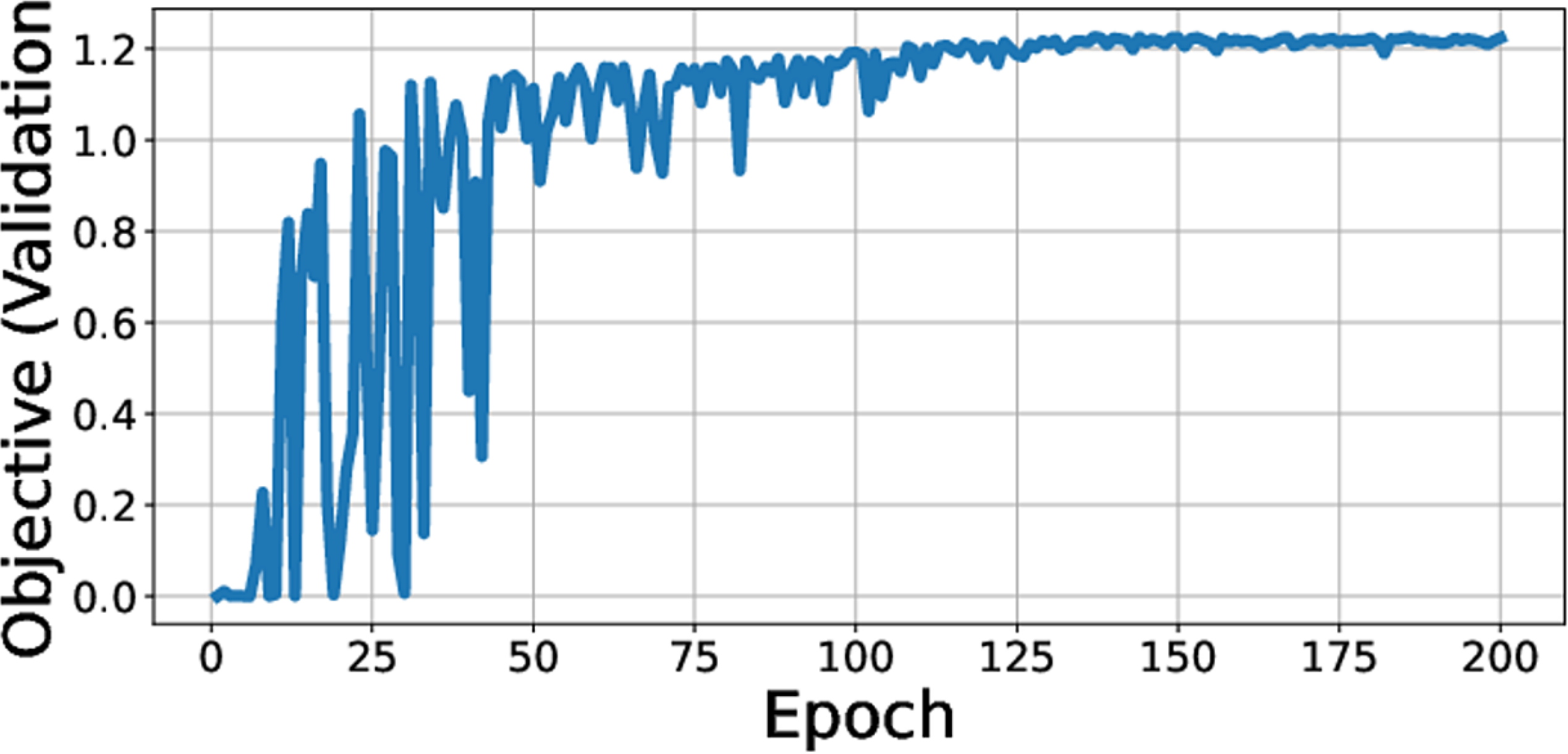}
        \caption{GS objective on validation set}
        \label{fig:valid_loss_epoch}
    \end{subfigure}
    \begin{subfigure}[b]{0.4\linewidth}
        \centering
        \includegraphics[width=\linewidth]{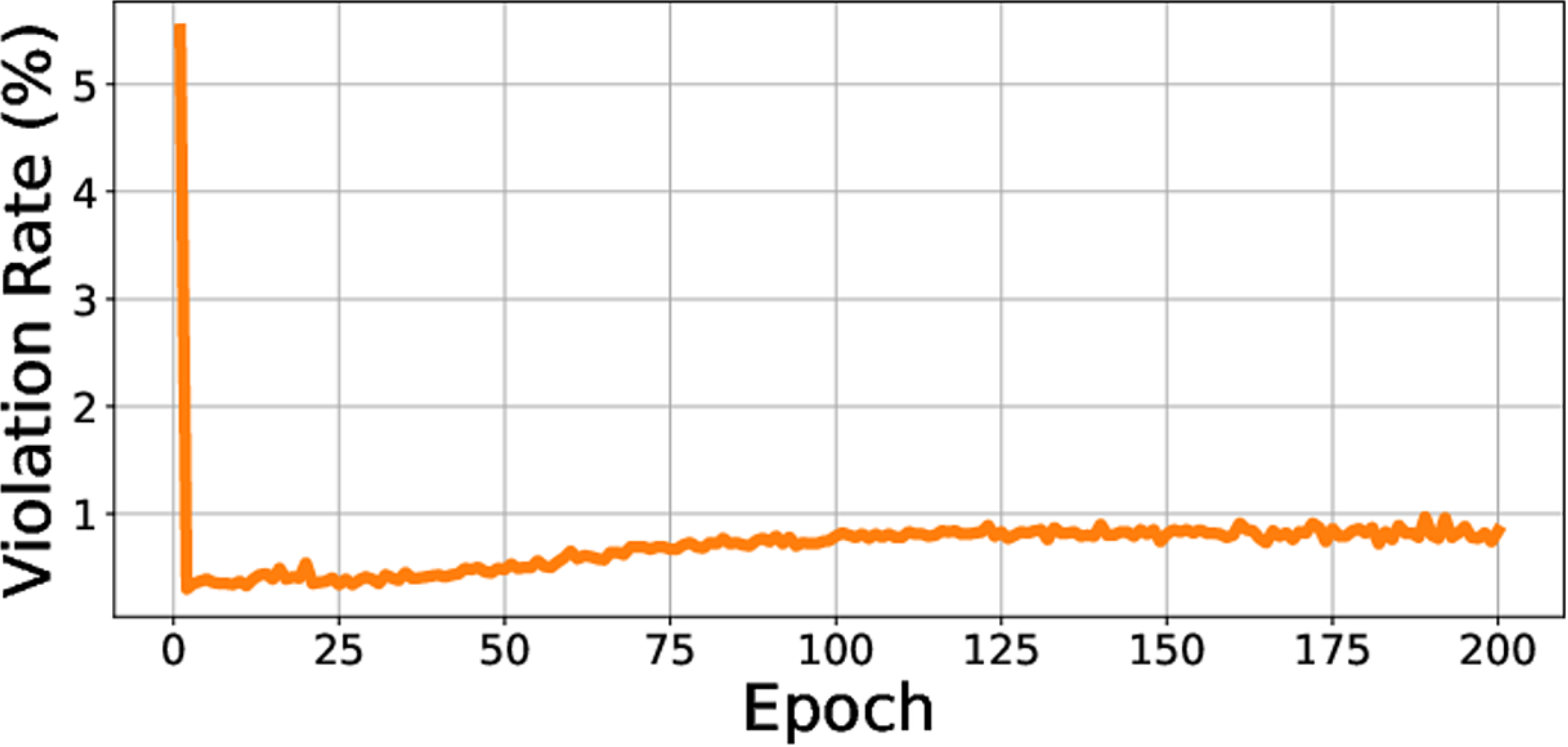}
        \caption{Constraint violation rate}
        \label{fig:constraint_epoch}
    \end{subfigure}
    \begin{subfigure}[b]{0.4\linewidth}
        \centering
        \includegraphics[width=\linewidth]{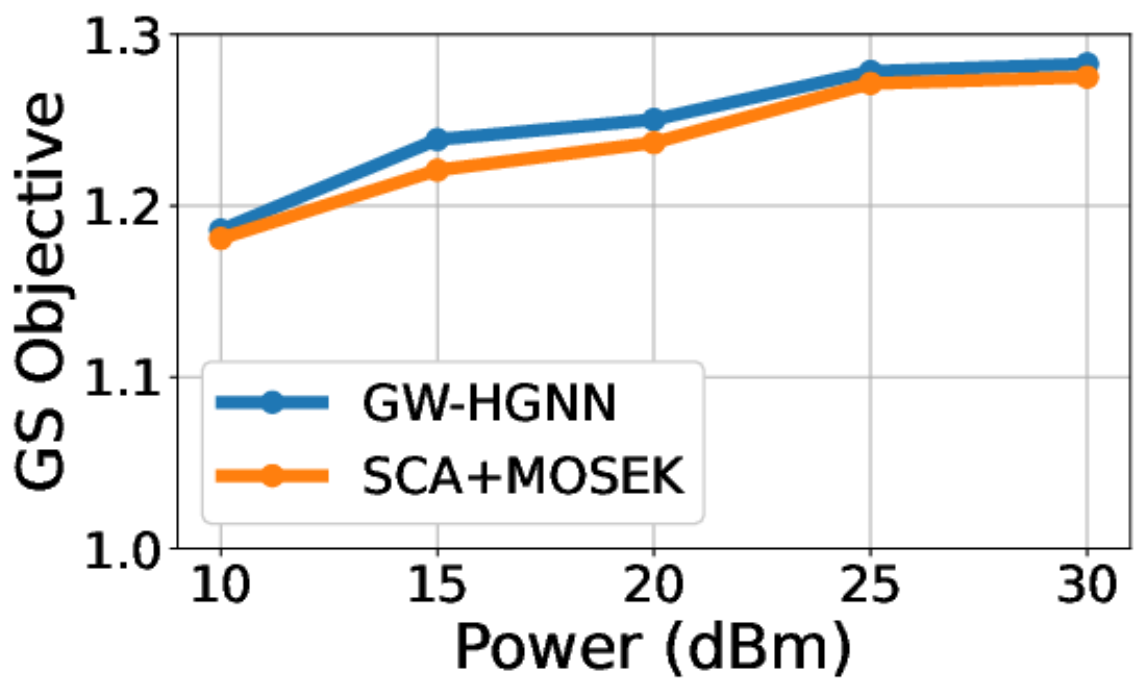}
        \caption{GS objective}
        \label{fig:sca_gnn_GSLoss}
    \end{subfigure}
    \begin{subfigure}[b]{0.4\linewidth}
        \centering
        \includegraphics[width=\linewidth]{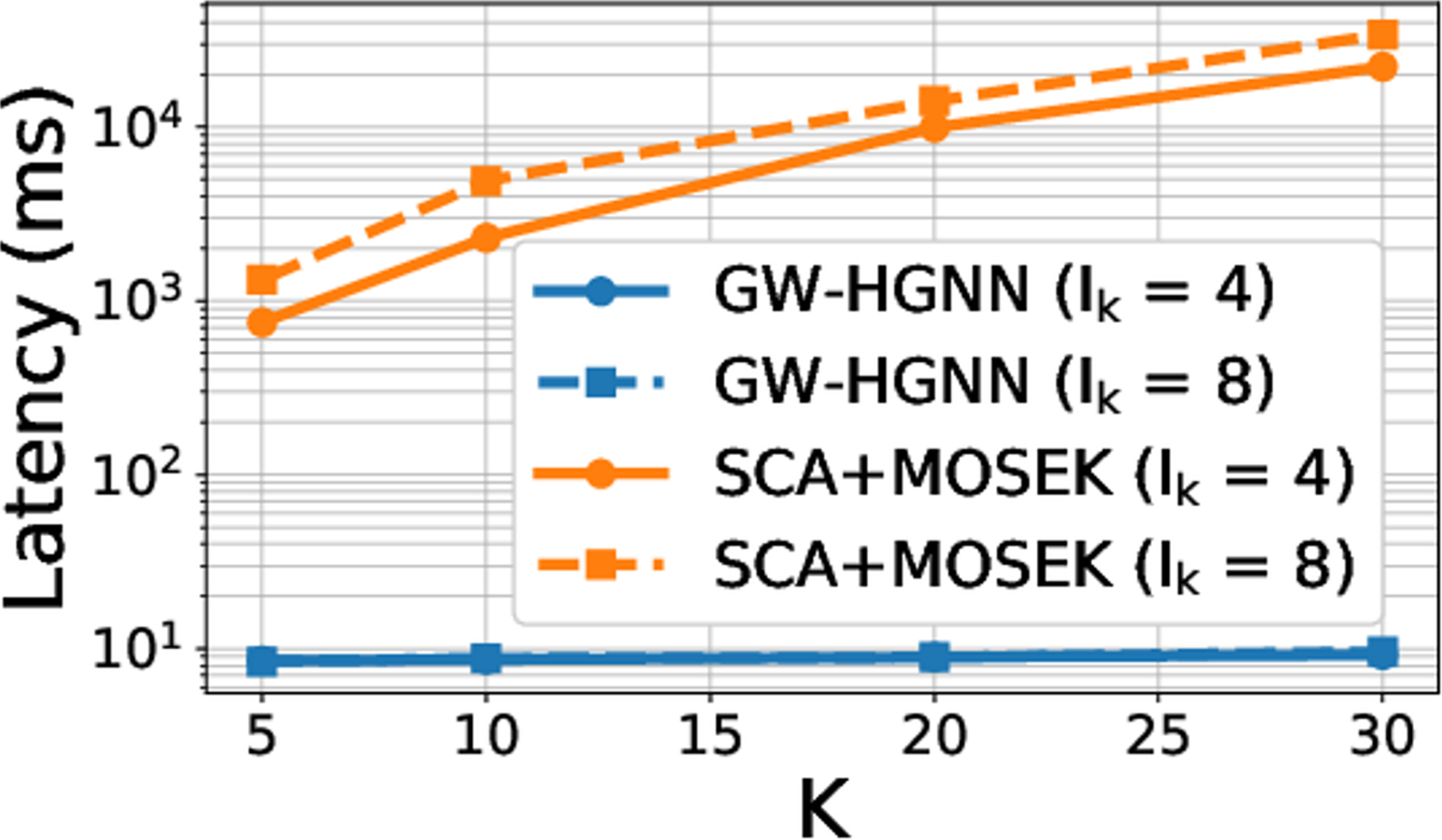}
        \caption{Average inference latency.}
        \label{fig:sca_gnn_latency}
    \end{subfigure}
    \caption{Performance and convergence analysis of the proposed framework. (a) GS objective on validation set. (b) Constraint violation rate. (c) Comparison of GS value. (d) Average inference latency.}
    \label{fig:combined_performance_analysis}
\end{figure*}

\begin{figure}[t]
    \centering
    \includegraphics[width=0.45\textwidth]{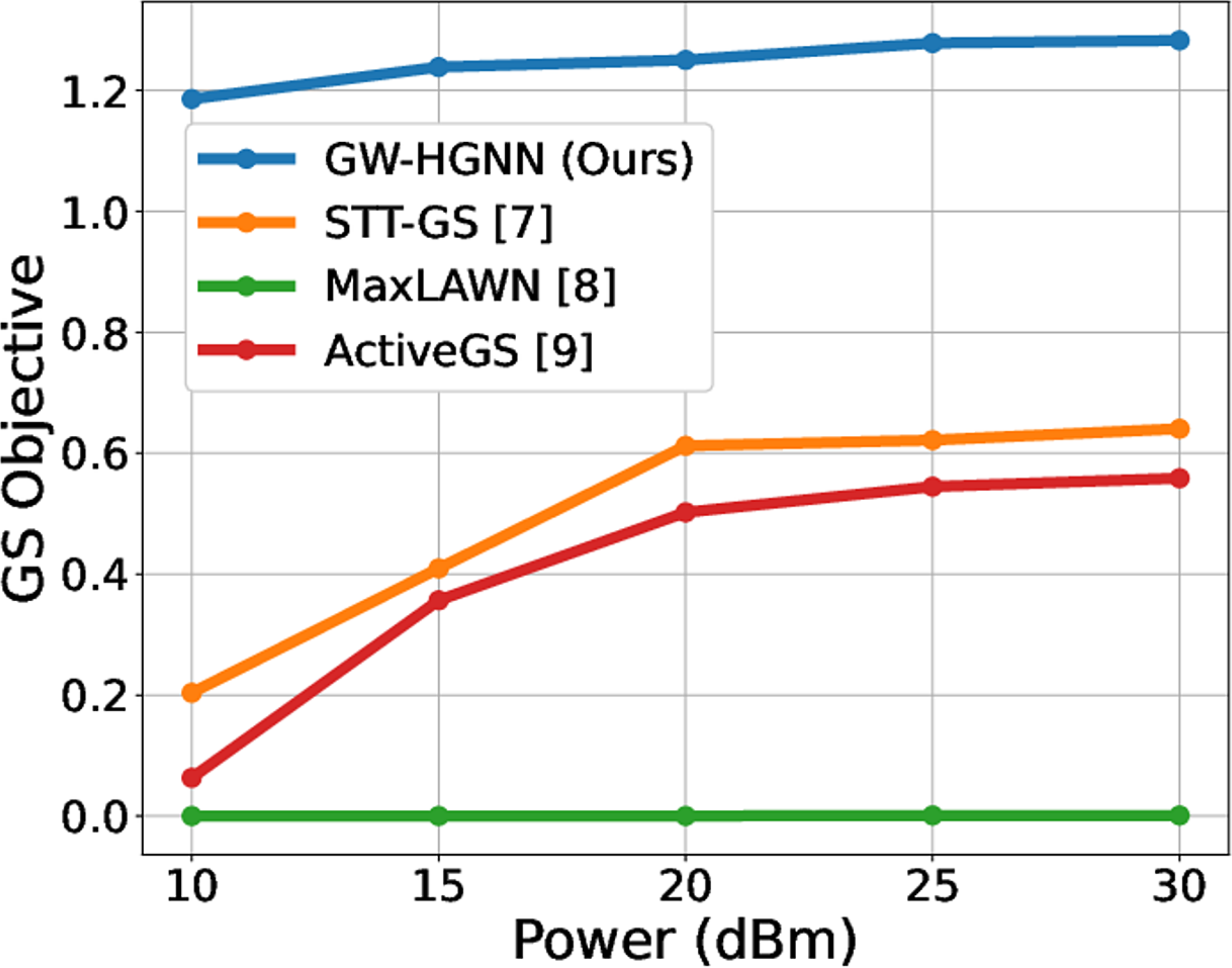}
    \caption{GS value versus power.}
    \label{fig:nongroup_objective}
\end{figure}

\begin{figure*}[t]
    \centering
    \begin{subfigure}[b]{0.37\linewidth}
        \centering
        \includegraphics[width=\linewidth]{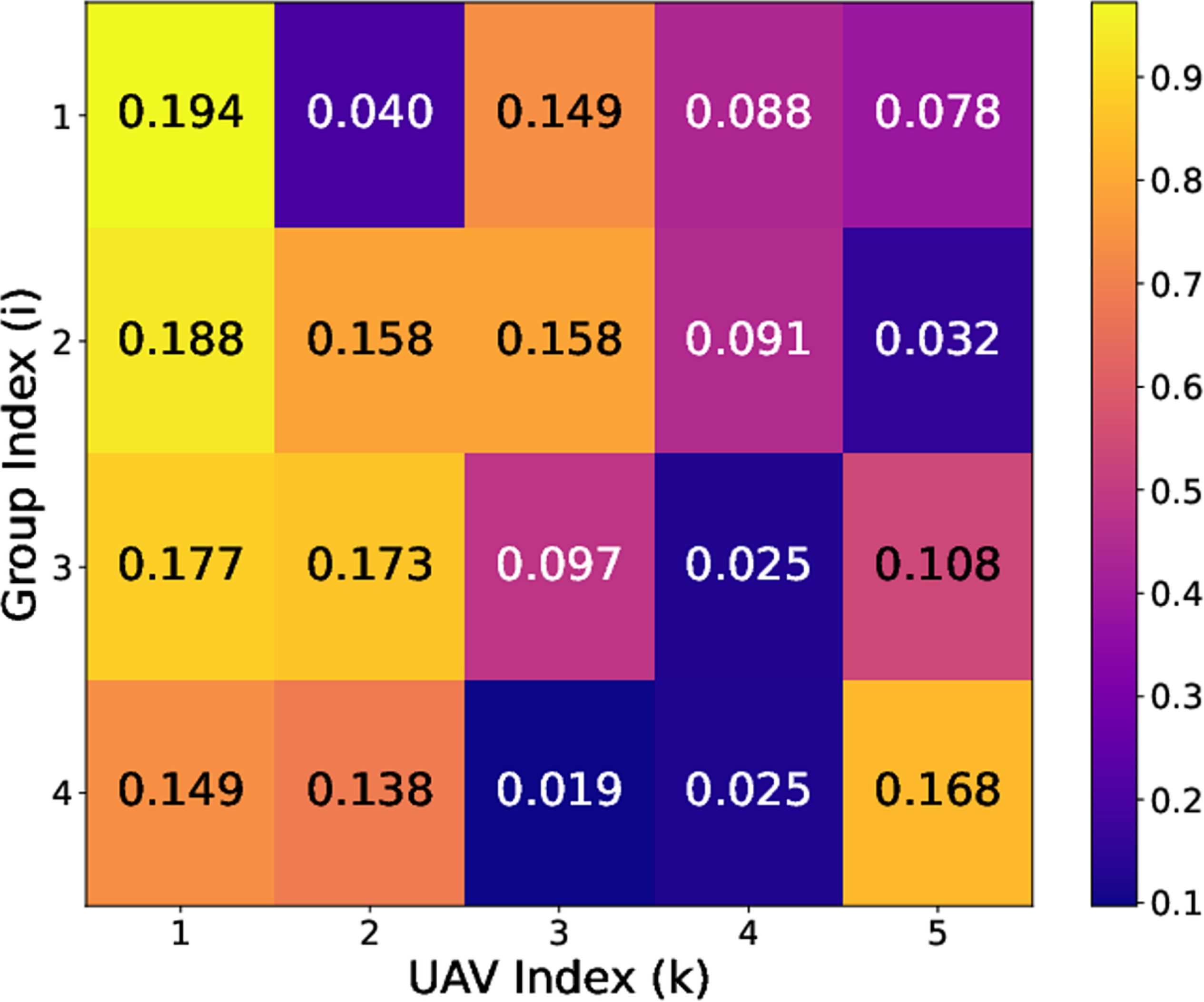}
        \caption{Visualization of $\pi_{ki}$.}
        \label{fig:compati_visual}
    \end{subfigure}
    \begin{subfigure}[b]{0.6\linewidth}
        \centering
        \includegraphics[width=\linewidth]{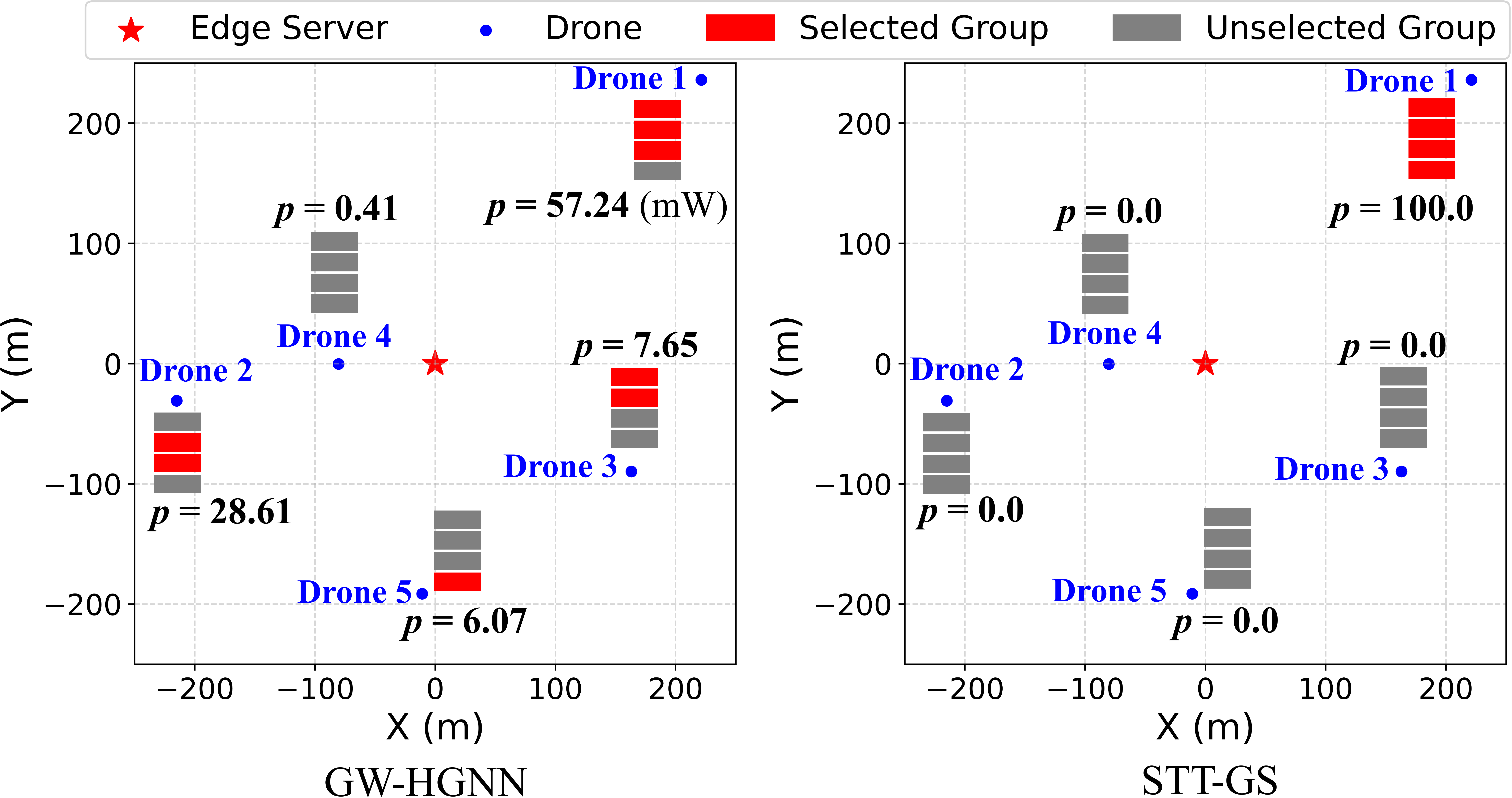}
        \caption{Visualization of group selection and power allocation.}
        \label{fig:Pos_Power_Group_Visual}
    \end{subfigure}
    \caption{Illustration of the case study.}
    \label{fig:overall_visual}
\end{figure*}

\section{Experiments}\label{section6}
We implement the LAGS system in Python using the 3DGS project~\cite{kerbl20233d}. The system is deployed on a Linux workstation with a single NVIDIA A6000 GPU.
We consider the case of $N=64$ and $K=5$, with drones randomly distributed in a $500 \times 500\,\text{m}^2$ area and the server located at the origin.
The channel $\mathbf{h}_{k}$ follows Rician fading \cite{Li2026STTGS,liu2026communication}:
\begin{equation} 
\mathbf{h}_k = \sqrt{h_0 \omega_k d_k^{-\alpha}} \left( \sqrt{\frac{K_{\text{Ric}}}{K_{\text{Ric}} + 1}} \mathbf{h}^{\text{LOS}}_{k} + \sqrt{\frac{1}{K_{\text{Ric}} + 1}} \mathbf{h}^{\text{NLOS}}_{k} \right),
\nonumber
\end{equation}
where $h_0=-30\,\text{dB}$, $\omega_k = -20\,\text{dB}$, $\alpha=2$, $K_{\text{Ric}}=10\mathrm{dB}$, and $d_k$ is the distance between drone $k$ and the server. 
The LoS component is modeled as
$\mathbf{h}_{k}^{\text{LoS}}=[1,e^{-j\pi\sin\theta_{k}},\dots,e^{-(N-1)j\pi\sin\theta_{k}}]^{T}$
with $\theta_{k}\in\mathcal{U}(-\pi,\pi)$, and NLoS component $\mathbf{h}_{k}^{\mathrm{NLOS}}\sim\mathcal{CN}(0,\mathbf{I}_{N})$. 
By default, MRC is adopted to generate $\{H_{kj}\}$.
The power budget is $P_{\text{sum}}=20\text{ dBm}$ and noise power is $\sigma^2 = -100\text{ dBm}$.
The total time budget is $T=50\text{ s}$\, and bandwidth is $B=3\, \text{MHz}$. 

We evaluate our method on the rubble-pixsfm dataset~\cite{turki2022mega}, which contains $1680$ real-world images collected by drones. {We reserve $168$ images for testing. For the remaining images, we adopt a trajectory-based viewpoint grouping strategy to partition them to groups. Specifically, the images are ordered according to their acquisition order along the UAV trajectory, which is consistent with the order of UAV image collection. The ordered image sequence of each drone is then divided into $I_k=4$ consecutive groups with approximately equal sizes. Therefore, the total number of IGs is $20$.}
The initial GS model $\mathcal{S}'$ is trained on $20\%$ data randomly sampled from the $20$ IGs.

The {GW-HGNN} employs $6$ message passing layers with hidden dimensions $[32, 64, 128,256,128,64]$. The model is trained on $51200$ samples for $200$ epochs with learning rate $10^{-4}$, batch size $128$, regularization coefficient $\psi=0.1$, and the Lagrange multipliers update step size $\tau=10^{-3}$. Validation and test set each contains $1024$ samples\footnote{We collect real data generated by GS models and assume the underlying distribution remains unchanged. If offline data collection is infeasible, online training is also viable due to the unsupervised manner of the training process.}.

We first validate the convergence behavior of the proposed GW-HGNN. 
Fig.~\ref{fig:valid_loss_epoch} shows the average GS value in \eqref{eq:op_with_prediction_loss} on validation dataset versus the number of epochs. 
GW-HGNN converges to a stable value after $100$ epochs. 
Fig.~\ref{fig:constraint_epoch} plots the constraint violation rate, defined as the fraction of drones whose scheduled data load exceeds their uplink capacity.
GW-HGNN remains below $1\%$ after a few epochs. 
This demonstrates that the Lagrangian-based learning effectively enforces communication constraint satisfaction. 

To further assess HGNN’s ability to solve $\sf{P}$, we compare it against a high-quality successive convex approximation (SCA) baseline that iteratively solves $\sf{P}$ by transforming it into a sequence of mixed-integer convex programs, each solved using the MOSEK solver~\cite{diamond2016cvxpy}. As shown in Fig.~\ref{fig:sca_gnn_GSLoss} and Fig.~\ref{fig:sca_gnn_latency}, GW-HGNN achieves GS {objective} close to SCA+MOSEK across all power budgets, but requires ultra-low inference latency ($<10 \text{ ms}$)\footnote{{The reported inference latency of GW-HGNN includes the network forward pass, thresholding, and the constraints enforcing procedure.}}. In contrast to SCA+MOSEK whose inference time grows rapidly with problem scale, our method exhibits scale-invariant latency thanks to its efficient feed-forward architecture.

\begin{table}[!t]
\caption{Comparison of Rendering Performance}
\label{tab:metric_table_non_group}
\centering
\scalebox{0.72}{
\begin{tabular}{l c c c }
    \toprule
    Method & {SSIM$\uparrow$} & {PSNR$\uparrow$}& {LPIPS$\downarrow$}   \\
    \midrule
    STT-GS \cite{Li2026STTGS} & 0.6622 & 20.61 & 0.3159 \\
MaxLAWN~\cite{ye2025integrated} & 0.6496 & 19.91 & 0.3375 \\
    ActiveGS~\cite{jin2025activegs} & 0.6622 & 20.61 & 0.3159 \\
    GW-HGNN (Ours) & 
        \cellcolor{pink!30}\textbf{0.7460} ($+13.2\%$) &
        \cellcolor{pink!30}\textbf{22.96} ($+11.4\%$) &
        \cellcolor{pink!30}\textbf{0.2687} ($-14.8\%$) \\
    \bottomrule
    \multicolumn{4}{l}{
    Note: Performance gain is computed against the STT-GS scheme.}
\end{tabular}
}
\end{table}

Next, we compare the performance of GW-HGNN against the following benchmarks: 1) \textbf{STT-GS} \cite{Li2026STTGS}: Cross-layer GS optimization without image partitioning; 2) \textbf{MaxLAWN}: Maximize sum rate of LAWN, which can be regarded as a simplified version of DeepLSC~\cite{ye2025integrated}\footnote{We adopt the objective function and power constraint in \cite{ye2025integrated} and ignore the trajectory and sensing constraints therein.}; 3) \textbf{ActiveGS}~\cite{jin2025activegs}\footnote{We assume perfect GS evaluation using ground truth images.  
}: Select data solely based on GS loss while ignoring channel conditions.

\begin{figure*}[t] 
    \centering
    \includegraphics[width=1\textwidth]{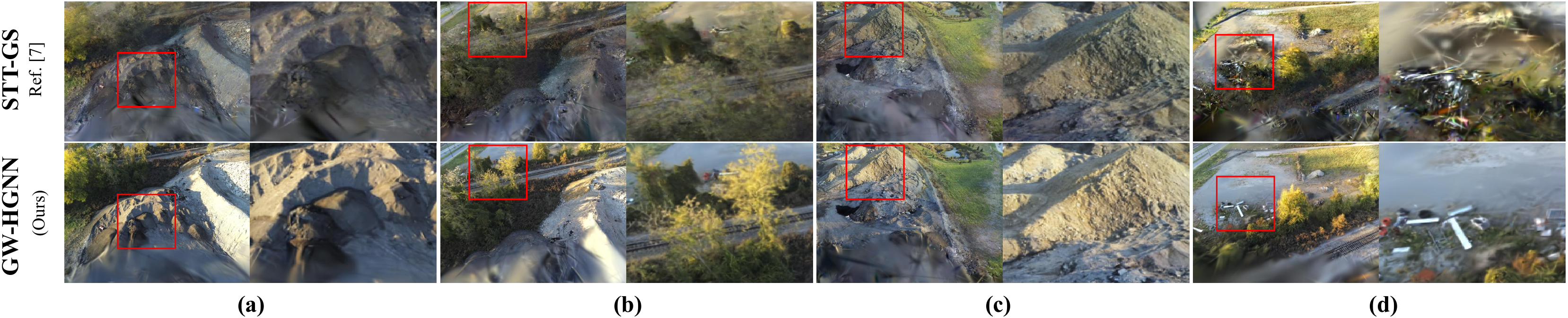}
    \caption{Visualization of rendered images for GW-HGNN and STT-GS schemes.}
    \label{fig:Render_visual}
\end{figure*}

Fig.~\ref{fig:nongroup_objective} plots the average GS objective~\eqref{eq:op_with_prediction_loss} over the test set across different power budgets. Our method outperforms STT-GS by up to $50\%$ in terms of the achievable GS objective, underscoring the advantage of explicitly modeling intra-drone viewpoint heterogeneity through GW scheduling. 
To provide further insight, we present a case study in Fig.~\ref{fig:compati_visual} and Fig.~\ref{fig:Pos_Power_Group_Visual}, which visualize $\pi_{ki}$, drone positions, power allocation, and group selection. The associated GS rendering metrics SSIM, PSNR, and LPIPS are compared in Table.~\ref{tab:metric_table_non_group}. 
We observe that STT-GS concentrates all power on drone $1$ with the highest $\pi_{ki}$. 
However, serving this drone has exhausted resources, since its hovering position is far from the server.
In contrast, GW-HGNN selects $3/4$ images from drone $1$, and trades the remaining resources to gather images from nearby drones $(2,3,5)$. 
This groupwise feature enables much more flexible scheduling.
Consequently, the proposed GW-HGNN achieves the best performance across all metrics. 
Compared to the second-best method STT-GS, our method increases the SSIM and PSNR by $13.1\%$ and $11.4\%$, respectively, and reduces the LPIPS by $14.8\%$. 
Fig.~\ref{fig:Render_visual} provides visualization of $4$ rendered images. 
The GW-HGNN generated images enjoy sharp outlines and realistic textures, whereas STT-GS generated images suffer from visible distortion and blurriness.

\begin{figure*}[t] 
    \centering
    \begin{subfigure}[b]{0.53\linewidth}
        \centering
        \includegraphics[width=0.98\linewidth]{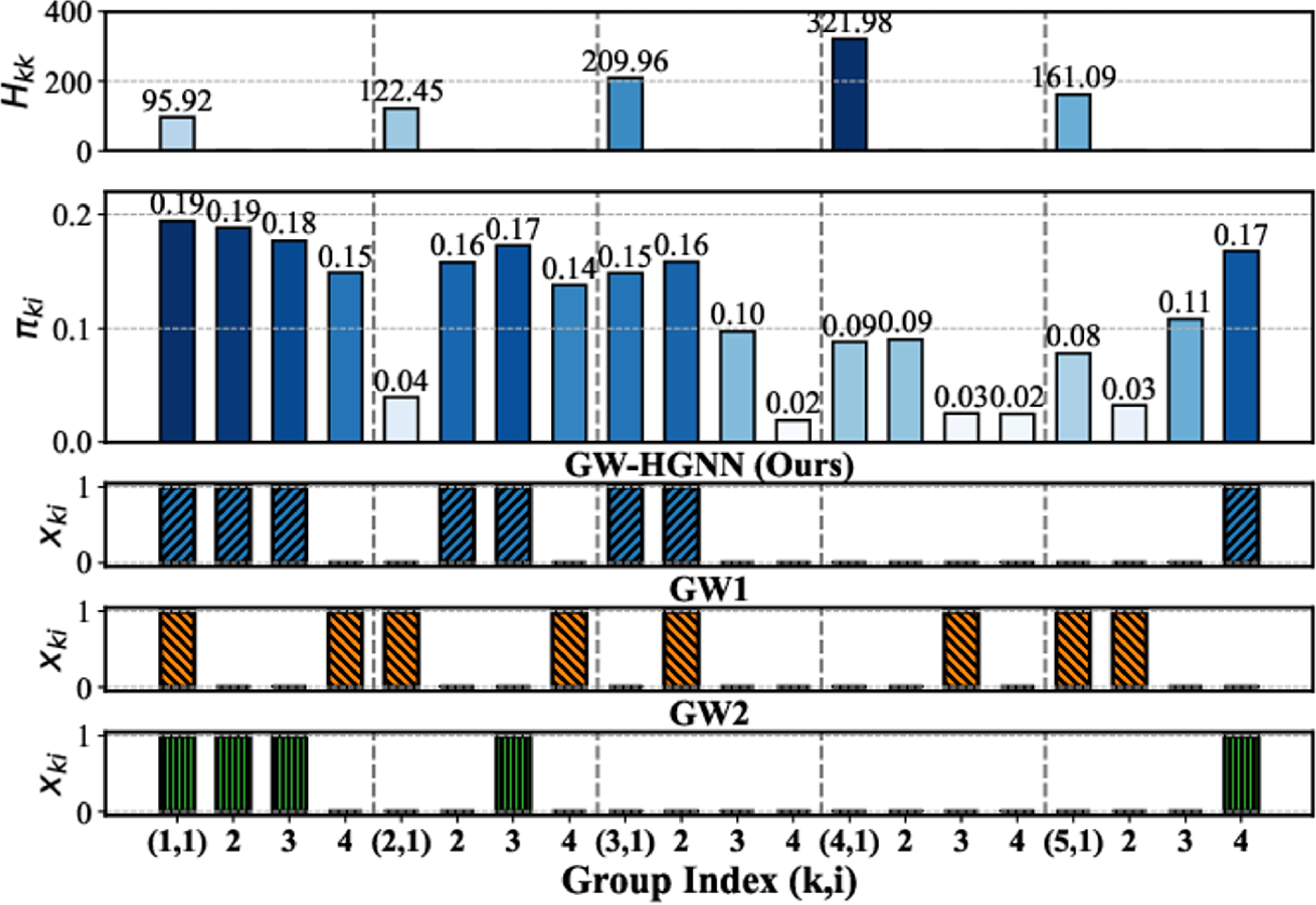}
        \caption{Visualization of group selection.}
        \label{fig:Group_Visualization}
    \end{subfigure}
    \begin{subfigure}[b]{0.46\linewidth}
        \centering
        \includegraphics[width=0.98\linewidth]{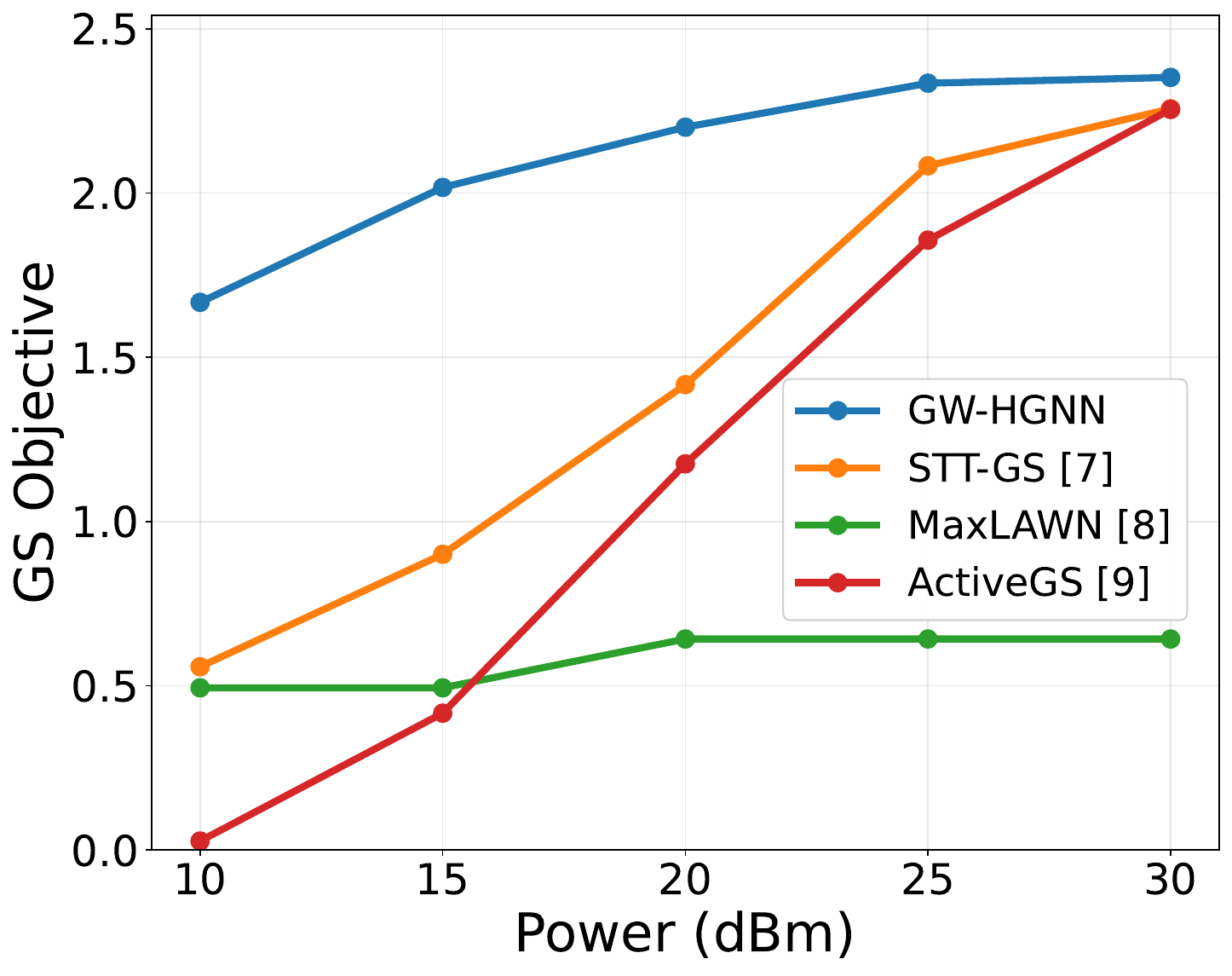}
        \caption{{GS value versus power (under the IRC combiner).}}
        \label{fig: gs_value_with_irc}
    \end{subfigure}
    \caption{Visualization of group selection and GS value comparison.}
\end{figure*}

Then, to assess the impact of cross-layer optimization, we conduct ablation studies comparing the GW-HGNN against two GW approaches: i.e., GW1 (which replaces the objective function in $\mathsf{P}$ with sum rate) and GW2 (which greedily selects the groups with the highest GS loss).
Results of group selection are visualized in Fig.~\ref{fig:Group_Visualization}, and quantitative rendering metrics (SSIM, PSNR, LPIPS) are reported in Table~\ref{tab:group_render_table}. 
It can be seen that GW1 fails in identifying the GS-valuable IGs while GW2 gathers the fewest images since it prioritizes the top-$5$ groups with highest $\pi_{ki}$.
Our method finds the best tradeoff between image quality and image quantity, achieving higher scores than GW1 and GW2 across all metrics. 
This confirms the necessity of leveraging both reconstruction and channel knowledge for LAGS.

\begin{table}[!t]
\caption{Comparison of Rendering Performance}
\label{tab:group_render_table}
\centering
\scalebox{0.85}{
\begin{tabular}{l c c c c }
    \toprule
    Method & {Objective$\uparrow$} & {SSIM$\uparrow$} & {PSNR$\uparrow$}& {LPIPS$\downarrow$}   \\
    \midrule
    GW1  & 0.7489 & 0.7101 & 21.95 & 0.2858 \\
    GW2  & 0.9001 & 0.7132 & 22.30 & 0.2874 \\
    GW-HGNN (Ours)  & \cellcolor{pink!30}\textbf{1.3656}  & \cellcolor{pink!30}\textbf{0.7460} 
                    & \cellcolor{pink!30}\textbf{22.96}  
                    & \cellcolor{pink!30}\textbf{0.2687} \\
    \bottomrule
\end{tabular}
}
\end{table}

{Finally, we further evaluate the proposed GW-HGNN under different uplink receiver. Specifically, we replace the MRC receiver with the IRC receiver while keeping the other experimental settings unchanged, and then evaluate the achievable GS objective under different power budgets. As shown in Fig.~1, GW-HGNN remains effective under IRC and consistently outperforms the baselines across all power values. This result demonstrates that the proposed framework is not tied to a specific uplink receiver design.}

\section{Conclusion}\label{section7}
This paper addressed the critical inefficiency of existing resource allocation schemes in LAGS systems, which overlook the image diversity introduced by varying viewpoints.
We proposed the GW-HGNN solution that learns the inter-drone and intra-drone relations with a heterogeneous graph. 
These relations successfully captured reconstruction and communication factors, enabling image group selection that maximizes the GS rendering quality.
Extensive experiments on real-world datasets validated the superiority of our approach. 

\bibliographystyle{IEEEtran}
\bibliography{ref}

\begin{IEEEbiography}
[{\includegraphics[width=1in, height=1.25in, clip, keepaspectratio]{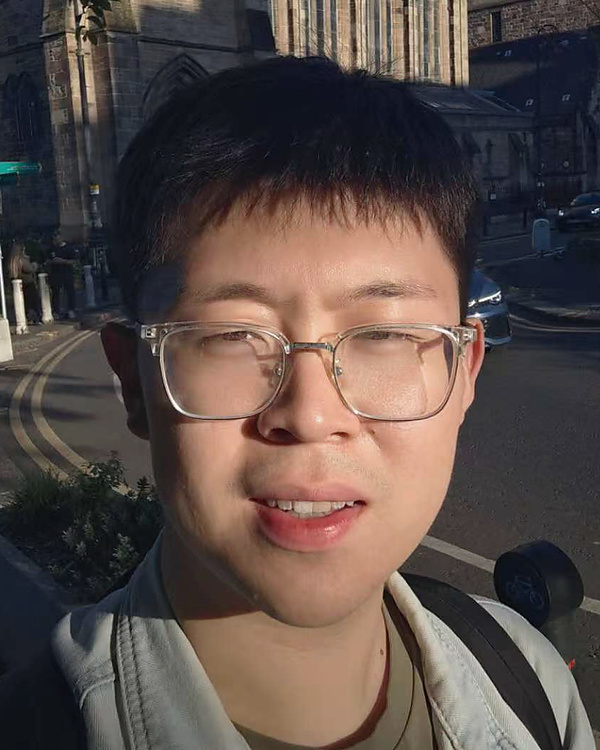}}]{Yikun Wang} (Graduate Student Member, IEEE) received his bachelor's degree from Beihang University (BUAA), Beijing, China, in 2024, majoring in communication engineering. He is currently pursuing the M.Phil. degree with the Department of Electrical and Computer Engineering, the University of Hong Kong, Hong Kong, China. His current research interests include 3D reconstruction.
\end{IEEEbiography}

\begin{IEEEbiography}
[{\includegraphics[width=1in, height=1.25in, clip, keepaspectratio]{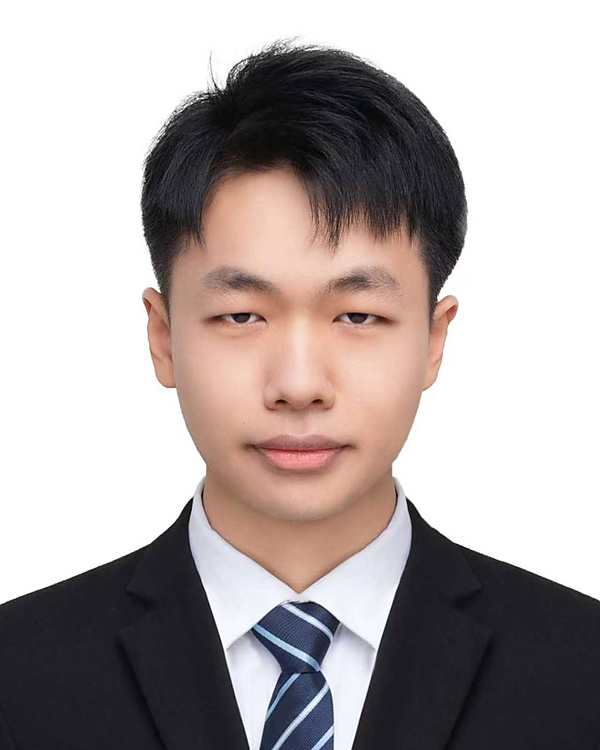}}]{Yujie Wan} (Graduate Student Member, IEEE) received the B.Eng. degree in software engineering from Chongqing University, Chongqing, China, in 2023. He is currently pursuing the M.Eng. degree in computer technology through the joint graduate program of the Southern University of Science and Technology and the Shenzhen Institutes of Advanced Technology, Chinese Academy of Sciences, Shenzhen, China. His research interests include robotic autonomous systems and 3D Gaussian splatting.
\end{IEEEbiography}

\begin{IEEEbiography}
[{\includegraphics[width=1in, height=1.25in, clip, keepaspectratio]{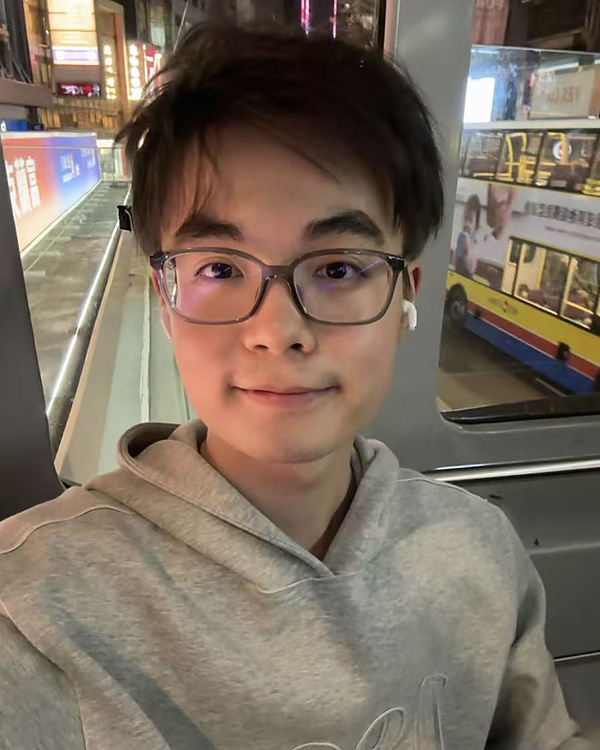}}]{Wei Zuo} (Graduate Student Member, IEEE) received his bachelor's degree from the Beijing Institute of Technology (BIT), Beijing, China, in 2024, majoring in automation. He is currently pursuing the M.Phil. degree with the Department of Electrical and Computer Engineering, the University of Hong Kong, Hong Kong, China. His current research interests include embodied navigation and spatiotemporal neural representation.
\end{IEEEbiography}

\begin{IEEEbiography}
[{\includegraphics[width=1in, height=1.25in, clip, keepaspectratio]{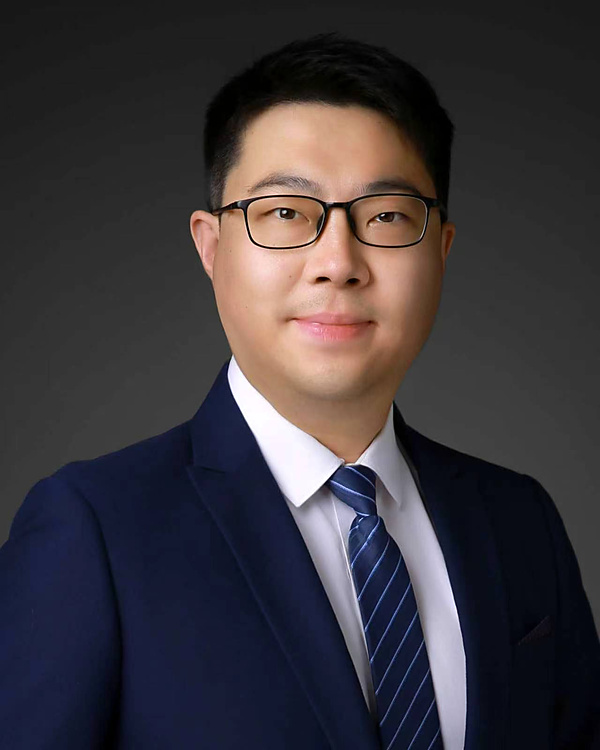}}]{Shuai Wang} (Senior Member, IEEE) received the B.Eng. and M.Eng. degrees from the Beijing University of Posts and Telecommunications (BUPT), Beijing, China, in 2011 and 2014, respectively, and the Ph.D. degree from the University of Hong Kong, Hong Kong, China, in 2018. He is currently a Professor with the Shenzhen Institutes of Advanced Technology (SIAT), Chinese Academy of Sciences, Shenzhen, China, where he leads the Intelligent Networked Vehicle Systems (INVS) Laboratory. His research interests include autonomous systems and wireless communications.
\end{IEEEbiography}

\begin{IEEEbiography}
[{\includegraphics[width=1in, height=1.25in, clip, keepaspectratio]{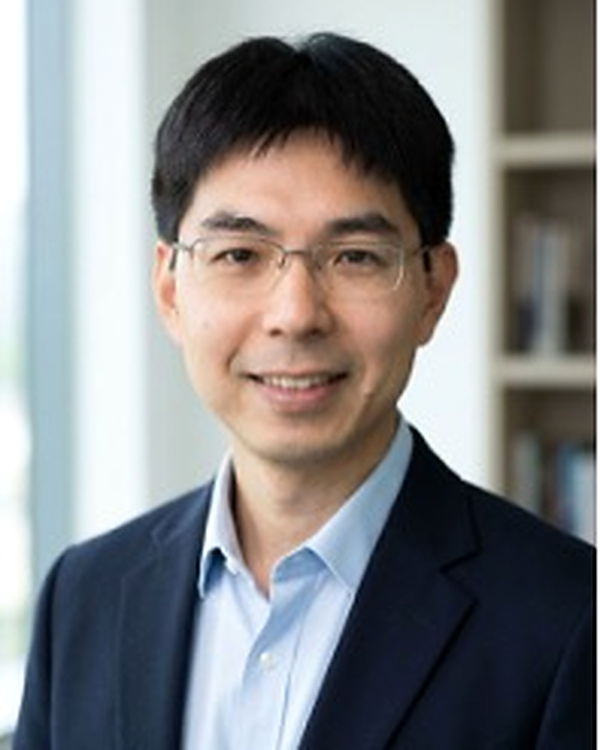}}]{Yik-Chung Wu} (Senior Member, IEEE) received the Ph.D. degree from Texas A\&M University, College Station, TX, USA, in 2005. After working as a Member of Technical Staff in Thomson Corporate Research, Princeton, NJ, USA, for one year, he has been with the University of Hong Kong since 2006, where he is currently an Associate Professor. He was a Visiting Scholar with Princeton University, in the summers of 2015 and 2017. His research interests are in the general areas of signal processing, machine learning, and communication systems, and in particular Bayesian inference, distributed algorithms, and large-scale optimization. Dr. Wu served as an Editor for IEEE Communications Letters and IEEE Transactions on Communications. He is currently a Senior Area Editor for IEEE Transactions on Signal Processing, an Associate Editor for IEEE Wireless Communications Letters, and an Editor for the Journal of Communications and Networks. He received four best paper awards in international conferences, with the most recent one from the IEEE International Conference on Communications (ICC) 2020. He was a Symposium Chair for many international conferences, including the IEEE International Conference on Communications (ICC) 2023 and IEEE Globecom 2025. He was elected the Best Editor of the Year 2023 in IEEE Wireless Communications Letters. He is an elected member of the IEEE Signal Processing Society SPCOM Technical Committee (2025--2026), and an IEEE Distinguished Lecturer (Vehicular Technology Society 2025 Class).
\end{IEEEbiography}

\begin{IEEEbiography}
[{\includegraphics[width=1in, height=1.25in, clip, keepaspectratio]{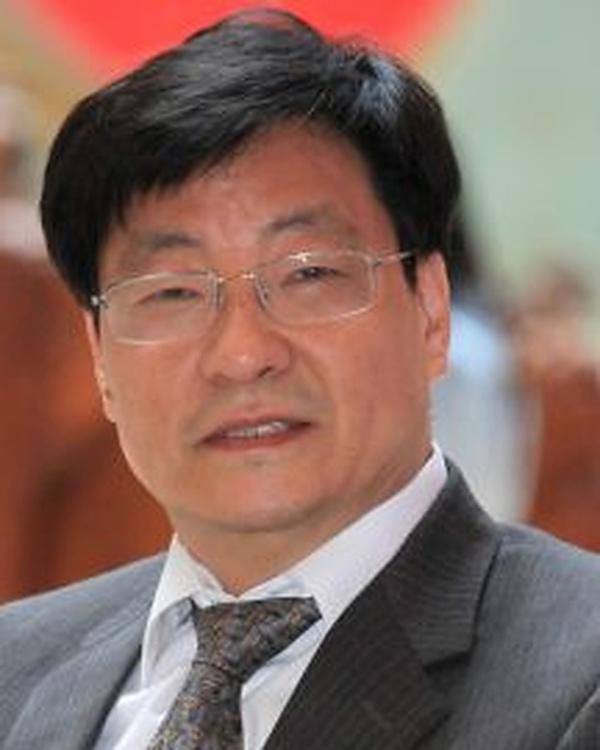}}]{Chengzhong Xu} (Fellow, IEEE) received the Ph.D. degree from the University of Hong Kong, Hong Kong, in 1993. He has held Faculty positions with Wayne State University, Detroit, MI, USA, and the Shenzhen Institutes of Advanced Technology, Chinese Academy of Sciences, Shenzhen, China. He is currently the Chair Professor of computer science with the University of Macau, Macau, China. He is also the Dean of Faculty of Information Science and Computing and the Director of Institute of AI and Brain Sciences.He has authored two research monographs and more than 600 journal and conference papers, garnering more than 28,000 citations, and an H-index of 86. His work has been cited in 370 international patents, and including 240 U.S. patents. His research interests include cloud and edge for AI, autonomous driving, and intelligent transportation. He is also a Co-Inventor of more than 200 PCT and China patents and a Co-Founder of the Shenzhen Institute of Beidou Applied Technology. His research has won him the Best Paper awards or nominations at conferences including SoCC 2021, HPCA 2013, HPDC 2013, and ICPP 2015. He formerly chaired the IEEE Technical Committee on Distributed Processing from 2015 to 2020. He is an IEEE Fellow for contributions to resource management in parallel and distributed computing.
\end{IEEEbiography}

\begin{IEEEbiography}
[{\includegraphics[width=1in, height=1.25in, clip, keepaspectratio]{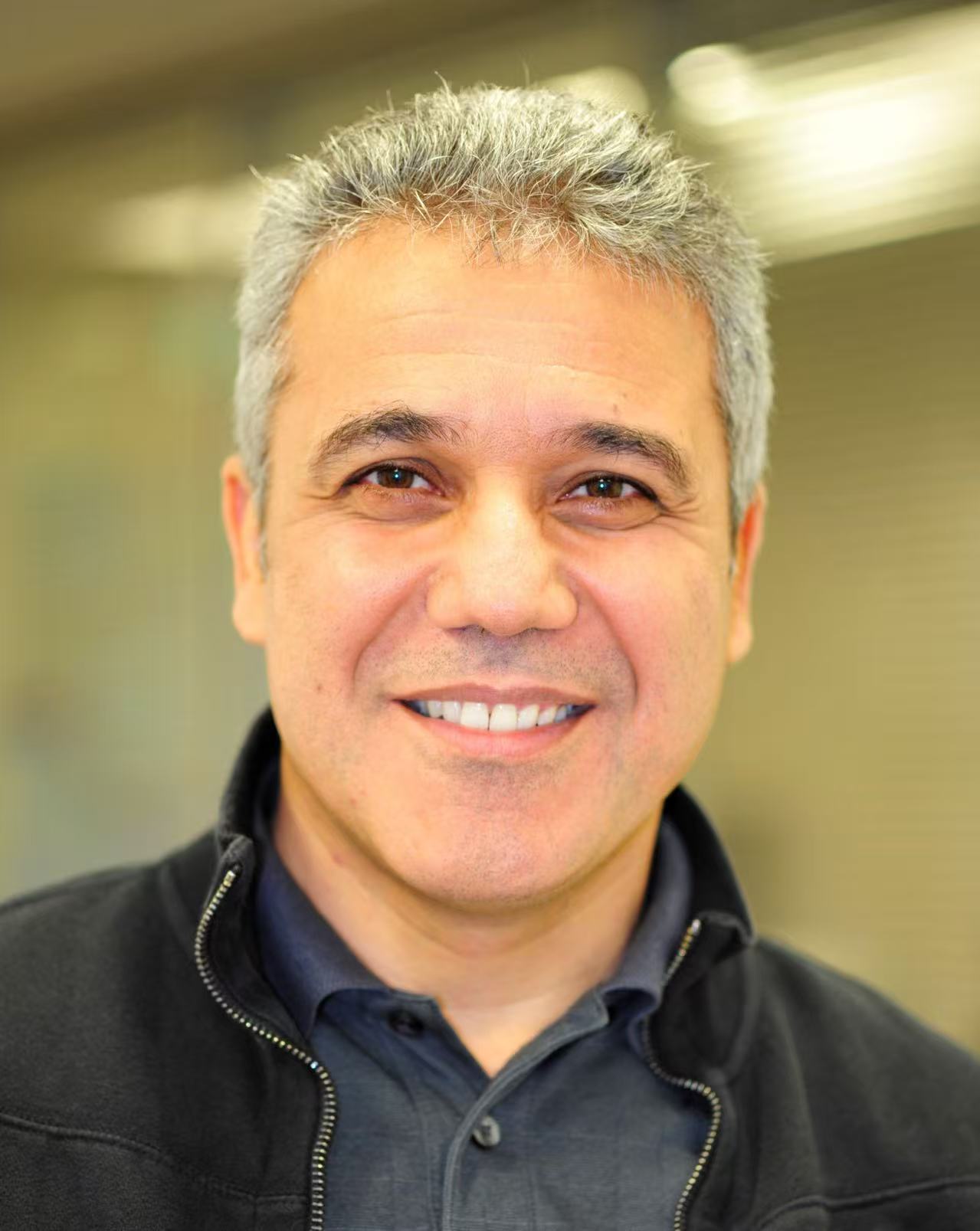}}]{Huseyin Arslan} (Fellow, IEEE) received the B.S. degree from the Middle East Technical University (METU), Ankara, Turkey, in 1992, and the M.S. and Ph.D. degrees from Southern Methodist University (SMU), Dallas, TX, USA, in 1994 and 1998, respectively. From January 1998 to August 2002, he was with the research group of Ericsson, where he was involved in several projects related to 2G and 3G wireless communication systems. Between August 2002 and August 2022, he was with the Electrical Engineering Department, University of South Florida, Tampa, FL, USA, where he was a Professor. In December 2013, he joined Istanbul Medipol University to found the Engineering College, where he has been working as the Dean of the School of Engineering and Natural Sciences. In addition, he has worked as a part-time consultant for various companies and institutions including Anritsu Company, Savronik Inc., and The Scientific and Technological Research Council of Turkey. Dr. Arslan served as the founding Chairman of the Board of Directors of ULAK Communication company, which is a Turkish telecom equipment provider. He was also a member of the TUBITAK Scientific Board. Between 2021 and 2024, he served as a Member of the Board of Directors for Turkcell, the biggest cellular operator in Turkey, while also operating in Ukraine, Belarus, and Cyprus.

\par Dr. Arslan conducts research in wireless systems, with an emphasis on the physical and medium access layers of communications. His current research interests are on 6G and beyond radio access technologies, physical layer security, interference management (avoidance, awareness, and cancellation), cognitive radio, multi-carrier wireless technologies (beyond OFDM), dynamic spectrum access, co-existence issues, non-terrestrial communications (High Altitude Platforms), and joint radar (sensing) and communication designs. Dr. Arslan has been collaborating extensively with key national and international industrial partners, and his research has generated significant interest in companies such as InterDigital, Anritsu, NTT DoCoMo, Raytheon, Honeywell, Aselsan, Vestel, Turkcell, and Keysight Technologies. Collaborations and feedback from industry partners have significantly influenced his research. In addition to his research activities, Dr. Arslan has also contributed to wireless communication education. He has integrated the outcomes of his research into education, which led him to develop a number of courses at the University of South Florida and Istanbul Medipol University. He has developed a unique ``Wireless Systems Laboratory'' course (funded by the National Science Foundation and Keysight Technologies) where he was able to teach not only the theory but also the practical aspects of wireless communication systems with the most contemporary test and measurement equipment.

\par Dr. Arslan has served as a General Chair, Technical Program Committee Chair, Session and Symposium Organizer, Workshop Chair, and Technical Program Committee Member in several IEEE conferences. He has also served as a member of the editorial board for IEEE Surveys and Tutorials, the IEEE Sensors Journal, IEEE Transactions on Communications, the IEEE Transactions on Cognitive Communications and Networking (TCCN), and several other scholarly journals by Elsevier, Hindawi, and Wiley Publishing.
\end{IEEEbiography}
\end{document}